\documentclass{article}

\usepackage{geometry}
\usepackage[utf8]{inputenc} 
\usepackage[T1]{fontenc}    
\usepackage{url}            
\usepackage{booktabs}       
\usepackage{amsfonts}       
\usepackage{nicefrac}       
\usepackage{microtype}      
\usepackage{bm}
\usepackage{float}
\usepackage{times}
\usepackage{epsfig}
\usepackage{graphicx}
\usepackage{amsmath}
\usepackage{amssymb}
\usepackage{algorithm}
\usepackage{algpseudocode}
\usepackage{xcolor,colortbl}
\usepackage{url}

\definecolor{Gray0}{rgb}{0.9,0.9,0.9}
\definecolor{Gray1}{rgb}{0.7,0.7,0.7}
\definecolor{Blue1}{rgb}{0.5,0.5,1.0}
\definecolor{Blue12}{rgb}{0.55,0.55,1.0}
\definecolor{Blue2}{rgb}{0.6,0.6,1.0}
\definecolor{Blue23}{rgb}{0.65,0.65,1.0}
\definecolor{Blue3}{rgb}{0.7,0.7,1.0}
\definecolor{Blue34}{rgb}{0.75,0.75,1.0}
\definecolor{Blue4}{rgb}{0.8,0.8,1.0}
\definecolor{Blue45}{rgb}{0.85,0.85,1.0}
\definecolor{Blue5}{rgb}{0.9,0.9,1.0}

\title{Confidence-Weighted Local Expression Predictions for Occlusion Handling in Expression Recognition and Action Unit detection}

\author{Arnaud Dapogny$^1$\\
{\tt\small arnaud.dapogny@isir.upmc.fr}
\and
Kevin Bailly$^1$\\
{\tt\small kevin.bailly@isir.upmc.fr}
\and
S\'{e}verine Dubuisson$^1$\\
{\tt\small severine.dubuisson@isir.upmc.fr}\\\\
$^1$ Sorbonne Universit\'{e}s, UPMC Univ Paris 06, CNRS, ISIR UMR 7222\\ 4 place Jussieu 75005 Paris
}

\begin{document}

\date{}
\maketitle

\begin{abstract}
  Fully-Automatic Facial Expression Recognition (FER) from still images is a challenging task as it involves handling large interpersonal morphological differences, and as partial occlusions can occasionally happen. Furthermore, labelling expressions is a time-consuming process that is prone to subjectivity, thus the variability may not be fully covered by the training data. In this work, we propose to train Random Forests upon spatially defined local subspaces of the face. The output local predictions form a categorical expression-driven high-level representation that we call Local Expression Predictions (LEPs). LEPs can be combined to describe categorical facial expressions as well as Action Units (AUs). Furthermore, LEPs can be weighted by confidence scores provided by an autoencoder network. Such network is trained to locally capture the manifold of the non-occluded training data in a hierarchical way. Extensive experiments show that the proposed LEP representation yields high descriptive power for categorical expressions and AU occurrence prediction, and leads to interesting perspectives towards the design of occlusion-robust and confidence-aware FER systems.
\end{abstract}

\section*{Introduction}\label{intro}

Automatic Facial Expression Recognition (FER) from still images is an ongoing research field which is key to many human-computer applications, such as consumer robotics or social monitoring. To address these problems, a lot of emphasis has been put by the psychological community in order to define models that are both accurate and exhaustive enough to describe facial expressions.

Perhaps one of the most long-standing model for describing the expressions is the discrete categorization proposed by Paul Ekman within his cross-cultural studies \cite{ekman1971constants}, in which he introduced six universally recognized basic expressions (\textit{happiness}, \textit{anger}, \textit{sadness}, \textit{fear}, \textit{disgust} and \textit{surprise}). Along with a \textit{neutral} state, this has been used as an underlying expression model for most attempts at developing a  prototypical expression recognition system \cite{lucey2010extended}, \cite{yin2008high}. However, this model faces limitations for dealing with spontaneous facial expressions \cite{zhang2014bp4d}, as many of our daily affective behaviors may not be translated in terms of prototypical emotions. Nevertheless, the annotation process is rather intuitive, thus there exists a large corpus of labelled data.

Another approach is the continuous affect representation \cite{greenwald1989affective} that consists in projecting expressions onto a restricted number of latent dimensions. A popular example of such model is the valence/activation (relaxed \textit{vs.} aroused)/power (feeling of control)/expectancy (anticipation) model. It is often simplified as a two-dimensional valence-activation representation. However, using such a low-dimensional embedding of facial expressions can cause the loss of information. Indeed, expressions such as \textit{surprise} are not represented correctly whereas others can not be separated well (\textit{fear} \textit{vs.} \textit{anger}). Last but not least, the annotation process is less intuitive than with the categorical representation.

Finally, an alternative facial expression model is the Facial Action Coding System (FACS) \cite{ekman1977facial}. It consists in describing facial expressions as a combination of $44$ facial muscle activations that are refered to as Action Units (AUs). AUs is a face representation that may be less subject to interpretation. It can theoretically be used in accordance with the so-called Emotional FACS (EMFACS) rules in order to describe a broader range of spontaneous expressions. However, the main drawback of the FACS-coding approach is that the annotation tends to be a time-consuming process. Furthermore, FACS coders have to be highly trained, hence limiting the quantity of available data.

In the meantime, as stated in \cite{zeng2009survey}, FER from still images is a challenging task as there may exist large variability in the morphology or in the expressiveness of different persons. Furthermore, countless configurations of partial occlusion can occasionally happen (e.g. with hand or accessories). As a result, this variability cannot be fully covered by restricted amounts of available training data. For those reasons, this paper introduces a new categorical expression-driven representation that we call Local Expression Predictions (LEPs). LEPs can be learned efficiently on the available expression datasets, and can serve multiple purposes such as occlusion handling for (global) categorical expression recognition, as well as confidence-aware AU detection.

\section{Related work}\label{related}

In this section we review recent approaches covering FER, with an emphasis on methods addressing the problem of partial occlusions. We also describe methods for AU detection.

\subsection{Occlusion handling in categorical FER}\label{categorical}

Most recent approaches covering FER from still images work in controlled conditions, on a frontal view and lab-recorded environments \cite{lucey2010extended,yin2008high}. Shan \textit{et al.} \cite{shan2009facial} evaluated the recognition accuracy of Local Binary Pattern (LBP) features. Zhong \textit{et al.} \cite{zhong2012learning} proposed to learn active facial patches that are relevant for FER. Zhao \textit{et al.} \cite{zhao2014unified} designed a unified multitask framework for simultaneously performing facial alignment, head pose estimation and FER. Such approaches showed satisfying results in constrained scenarios, but they can face difficulties on more challenging benchmarks \cite{dhall2011static}.

Eleftheriadis \textit{et al.} \cite{eleftheriadis2015discriminative} used disciminative shared Gaussian processes to perform pose-invariant FER. Liu \textit{et al.} \cite{liu2015inspired} introduced a deep neural network that learns local features relevant for Action Unit prediction, and use it as an intermediate representation for categorical FER. The authors also studied the use of unlabelled data \cite{liu2013enhancing} to regularize the network training, further enhancing its predictive capacities for FER in the wild. However, none of these approaches explicitly addresses the problem of facial occlusions that are likely to happen in such unconstrained cases.

Kotsia \textit{et al.} \cite{kotsia2008analysis} studied the impact of human perception of facial expressions under partial occlusions, and the predictive capacities of automated systems thereof. Cotter \textit{et al.} \cite{cotter2010sparse} used sparse decomposition to perform FER on corrupted images. Ghiasi \textit{et al.} \cite{ghiasi2014occlusion} use a discriminative approach for facial feature point alignment under partial occlusions. Those approaches rely on explicitly incorporating synthetic occluded data in the training process, and thus struggle to deal with realistic, unpredicted occluding patterns. Zhang \textit{et al.} \cite{zhang2014random} trained classifiers upon random Gabor-based templates. They evaluated their algorithms on synthetically occluded face images, showing that their approach leads to a better recognition rate when the same occluded examples are used for training and testing. Should this not be the case, unpredicted mouth/eye occlusions still lead to a significant loss of performance. Huang \textit{et al.} \cite{huang2012towards} proposed to automatically detect the occluded regions using sparse decomposition residuals. However, the proposed approach may not be flexible enough, as the occlusion detection only outputs binary decisions, and as the face is explicitly divided into only three subparts (eyes, nose and mouth). This limits the capacities of the method to deal with unpredicted forms of occlusion. Finally, another approach consists in learning generative models of non-occluded faces, as it was done by Ranzato \textit{et al.} \cite{ranzato2011deep}. When testing on a partially occluded face image, the occluded parts can be generated back and expression recognition can be performed. The pitfall of such an approach is that training can be computationally expensive and does not allow the use of heterogeneous features (e.g. geometric/appearance descriptors).

\subsection{Action Unit detection}\label{relatedAU}

\begin{figure*}[ht]
\centering
\includegraphics[width=\textwidth]{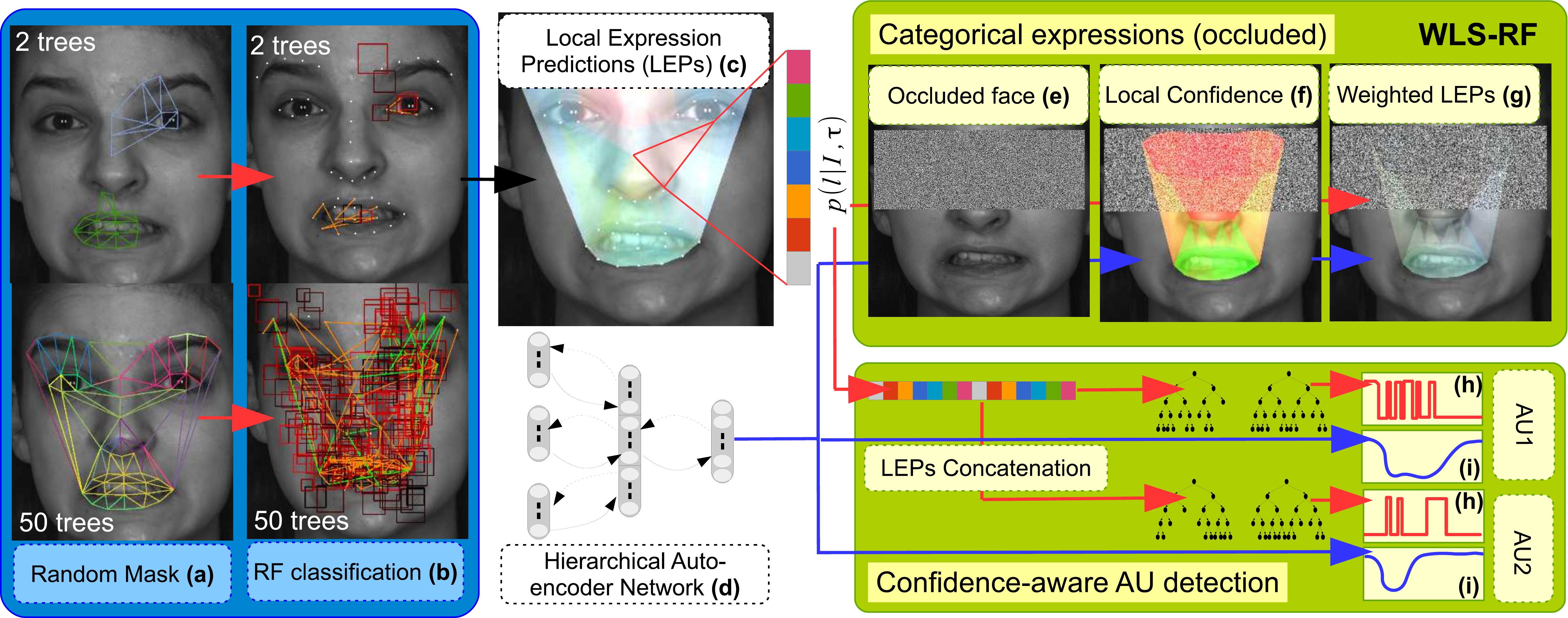}
\caption{LEPs and applications to categorical expression recognition, occlusion handling in FER and AU detection. Randomized trees are trained upon local subspaces generated under the form of random facial masks \textbf{(a)}, on which binary feature candidates are generated and selected \textbf{(b)}. The local predictions outputted by the trees can be aggregated into categorical expression-driven high-level LEP representations \textbf{(c)}. Given an occluded face image \textbf{(e)}, an occlusion-robust categorical expression prediction can be outputted by weighting LEPs with confidence scores \textbf{(f)} given by a hierarchical autoencoder network \textbf{(d)}. Furthermore, LEP features can be used to predict AU occurrence \textbf{(h)}, for which an AU-specific confidence measurement can be provided \textbf{(i)}. Best viewed in color.}
\label{flowchart}
\end{figure*}

AU detection is traditionally performed by applying binary classification upon high-dimensional, low-level image descriptors such as LBP or Local Phase Quantification (LPQ) features \cite{jiang2011action}. Senechal \textit{et al.} proposes to embed heterogeneous (geometric/appearance) features within a multi-kernel SVM framework \cite{senechal2012facial}. These features can be extended to spatio-temporal volumes with the Three Orthogonal Planes (TOP) paradigm, as proposed in \cite{zhang2014bp4d} for LBP-TOP and \cite{jiang2011action} for LPQ-TOP. In the meantime, Chu \textit{et al.} \cite{chu2013selective} introduced a new learning algorithm that personalizes a generic classification framework by attenuating person-specific biases. Nicolle \textit{et al.} \cite{nicolle2015facial} proposed an AU intensity prediction by a novel multi-task formulation of the metric learning for kernel regression method. However, there seems to be a gap between low-level feature descriptors (e.g. LBP, LPQ, SIFT) and high-end learning algorithms applied to AU detection, that could be filled by learning representations from a large corpus of categorical expression-labelled examples.

Recently, Ruiz \textit{et al.} \cite{ruiz2015shtl} introduced a new framework where categorical expressions are learnt from prior knowledge between this visible task and a set of hidden tasks that correspond to AU detection predictions. Thus, they exploit relationships between those two tasks to learn AU detectors with (SHTL) or without (HTL) FACS-labelled training data, by explicitly combining AUs into categorical expressions \textit{a la} EMFACS. Conversely, we propose to describe AUs as a combination of Local Expression Predictions (LEPs). Furthermore, to the best of our knowledge, this is the first time that a confidence assessment is provided for AU detection.

\section{Method Overview}\label{overview}

In this work, we introduce a new Local Expression Prediction (LEP) representation that can be learned from data labelled with categorical expressions, as described in Figure \ref{flowchart}. During training, local subspaces are generated under the form of random facial masks (a), onto which binary candidate features can be selected (b) to train randomized trees. The local LEPs (c) outputted by local subspace Random Forests can be used for multiple purposes that are depicted below. Furthermore, we also introduce a hierarchical autoencoder network (d), which can be used to capture the local manifold of non-occluded faces around separate aligned feature points. When applied on a potentially occluded face image (e), the reconstruction error outputted by such a network provides a confidence measurement of how close a face region lies from the training data manifold (f), with high and low confidences depicted in green and red respectively. This local confidence measurement can be used to weight LEPs (g) in order to provide an occlusion-robust expression prediction (WLS-RF). Finally, LEPs can be used to predict AU occurrence (h). Once again, the autoencoder network can be used to provide AU-specific confidence measurements (i). Our contributions are thus the following:

(1) A method for training random trees upon spatially-defined local subspaces, which consists in generating random masks covering a specified fraction of the face. These local trees can be combined to produce high-level expression-driven representations that we refer to as LEPs, which can be used for categorical FER or AU prediction.

(2) A hierarchical autoencoder network for learning local non-occluded face manifolds, which can be used to provide local confidence measurements.

(3) The confidence measurements outputted by the autoencoder network can be used to enhance robustness to occlusions for categorical FER, as well as to assess confidence for AU detection.

The rest of the paper is organized as follows: in Section \ref{aec} we discuss the proposed autoencoder network architecture (Section \ref{networkarchi}) and how it is trained to capture the local manifold around facial feature points (Section \ref{trainaec}). Section \ref{local} describes how we learn the Local Expression Predictions via local subspace random forests (Section \ref{learning}) with heterogeneous binary feature candidates (Section \ref{features}). In particular, we explain in Section \ref{localtest} how those local representations can be effectively combined and weighted to produce occlusion-robust predictions, and in Section \ref{pautoau} how LEPs can be used for confidence assessment in AU detection. Finally, in Section \ref{exper} we show that our approach significantly improves the state-of-the-art for categorical FER on multiple datasets (described in Section \ref{datas}), both on the non-occluded (Section \ref{clean}) and occluded cases (Section \ref{corr}). We also demonstrate in Section \ref{AUsexp} the interest of our LEP representation for AU activation prediction and the relevance of the AU-specific confidence measurement. Finally, Section \ref{concl} provides a conclusion as well as a few perspectives raised in the paper.

\section{Manifold Learning of non-occluded faces with a hierarchical autoencoder network}\label{aec}

Given a number of aligned facial feature points that can be provided by an off-the-shelf alignment algorithm (such as the SDM tracker \cite{xiong2013supervised}), we use an autoencoder network to model the local face pattern manifold. This network will thus be used to provide a local confidence measurement that is used to weight LEPs for occlusion-robust FER.

\subsection{Network architecture}\label{networkarchi}

Autoencoders are a particular type of neural network that can be used for manifold learning. Compared with other approaches such as PCA \cite{jolliffe2002principal}, autoencoders offer the advantage to theoretically be able to model complex manifolds using non-linear encoding and regularization criteria, such as denoising \cite{vincent2010stacked} or contractive penalties \cite{rifai2011contractive}. As compared to manifold forests \cite{pei2013unsupervised}, autoencoders can be trained on high-dimensional features without falling into the pitfall of low-rank deficiency. Furthermore, they benefit from an efficient training using stochastic gradient descent, as well as the possibility of online fine-tuning for subject-specific calibration.

\begin{figure}[htbf]
\centering
\includegraphics[width=0.8\linewidth]{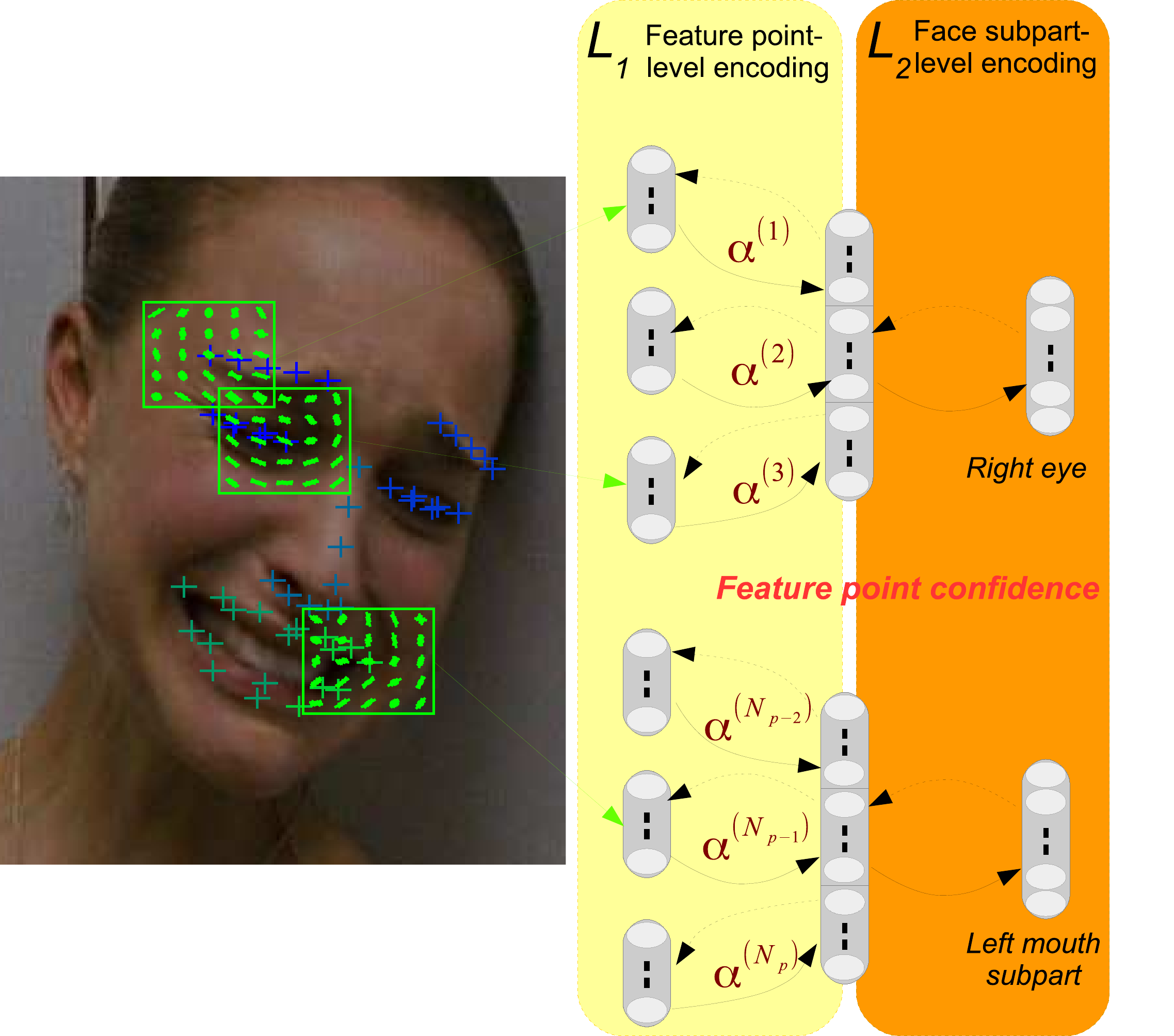}
\caption{Architecture of our hierarchical autoencoder network. The network is composed of 2 layers: the first one ($L_1$) captures the texture variations (HOG descriptors) around the separate aligned feature points. The second one ($L_2$) is defined over 5 face subparts, each of which embraces multiple points whose appearance variations are closely related. The network outputs a confidence score $\alpha^{(k)}$ for each of the $N_p$ feature points.}
\label{netarchi}
\end{figure}

As shown in Figure \ref{netarchi}, we use a 2-layer architecture. First, we extract Histograms of Oriented Gradients (HOG) within the neighbourhood of each feature point aligned on the face image $\mathcal{I}$. The choice of learning a manifold of HOG patterns rather than gray levels comes from the fact that HOG are used for both facial alignment and the LEP generation pipeline. Thus, the reconstruction error of these patterns provides a confidence measurement that is relevant for both tasks. In order to ensure fast HOG extraction, we use integral feature channels as introduced in \cite{dollar2009integral}. Horizontal and vertical gradients are computed on the image and used to generate $9$ feature maps. The first of these contains the gradient magnitude, and the $8$ remaining  correspond to a 8-bin quantization of the gradient orientation. Then, integral images are computed from these feature maps to output the $9$ feature channels. Also, storing the gradient magnitude within the first channel allows to normalize the histograms as in standard HOG implementations. Thus, HOG features can be computed very efficiently by using only $4$ access to previously stored integral channels (plus normalization). Note that those feature channels can be computed only once for the three steps of the pipeline.

The local descriptor $\mathbf{\Psi}^{(k)}$ for a specific feature point $k$ consists in the concatenation of gradient magnitudes and quantized orientation values in $5 \times 5$ cells around this feature point, with a total window size equal to a third of the inter-ocular distance. This descriptor of dimension $225$ then feeds the $N_p$ autoencoders (one per feature point) of the first layer ($L_1$) which are trained to reconstruct non-occluded patterns. Because occlusion of local patterns extracted at the feature point level are not independent (\textit{i.e.} a feature point close to an occluded area is more likely to be occluded itself), we employ a second layer ($L_2$) of autoencoders, that are trained to reconstruct non-occluded patterns of groups of encoded feature point descriptors. Those groups represent five face subparts (left and right eyes, nose, left and right parts of the mouth) from which the local patterns are closely related. Specifically, $L_1$ is composed of 125 units for each landmark. $L_2$ layer for a feature point group contains $65 \times N$ units ($\frac{1}{2}$ compression), where $N$ is 12,12,8,11 and 11 respectively for left/right eye, nose and left/right mouth areas.

\subsection{Training the network}\label{trainaec}

Autoencoders are generally trained in an unsupervised way, one layer at a time, by optimizing a reconstruction criterion. The input descriptor $\mathbf{\Psi}^{(k)}$ at feature point $k$ is first encoded via the $L_1$ encoding layer into an intermediate representation $\mathbf{y}^{(k)}=h^1(\mathbf{\Psi}^{(k)})$:

\begin{equation}\label{enc}
\mathbf{y}^{(k)} = \sigma \left(\bm{w}^{(k)}.\mathbf{\Psi}^{(k)} + \mathbf{b}^{(k)}\right)
\end{equation}

Where $\sigma$ is a sigmoid function, $\bm{w}^{(k)}$ and $\mathbf{b}^{(k)}$ are respectively the neuron weight matrix and bias vector of the $L_1$ neuron layer for feature point $k$. The output is then typically computed as the input reconstruction $\mathbf{\tilde{\Psi}}^{(k)}=g^1(\mathbf{y}^{(k)})$ using an affine decoder with tied input weights to reduce the number of parameters:

\begin{equation}\label{dec}
\mathbf{\tilde{\Psi}}^{(k)} = \bm{w}^{(k)T}.\mathbf{y}^{(k)} + \mathbf{c}^{(k)}
\end{equation}

Where $c^{(k)}$ is the decoder bias vector. Then the set of $K$ encoded descriptors $\{\mathbf{\tilde{\Psi}}^{(k)}\}_{k=1...K}$ associated to feature points ${k=1...K}$ that belong to the face subpart $m$ are concatenated to form the input $\mathbf{\xi}^{(m)}$ of the layer $L_2$ for that subpart. Once again, the input of the $L_2$ layer is successively encoded into a intermediate representation $\mathbf{z}^{(m)}=h^2(\mathbf{\xi}^{(m)})$ and decoded in the same way into a reconstructed version $\mathbf{\tilde{\xi}}^{(m)}=g^2(\mathbf{z}^{(m)})$:

\begin{equation}\label{enc2}
\mathbf{z}^{(m)} = \sigma \left({\bm{w'}^{(m)}.\mathbf{\xi}^{(m)}} + \mathbf{b'}^{(m)}\right)
\end{equation}

\begin{equation}\label{dec2}
\mathbf{\tilde{\xi}}^{(m)} = \bm{w'}^{(m)T}.\mathbf{z}^{(m)} + \mathbf{c'}^{(m)}
\end{equation}

Thus, each layer is trained separately using stochastic gradient descent and backpropagation. More specifically, the input descriptors for each layer are presented sequentially. For example, a forward pass through the $L_1$ layer provides a reconstructed version $\mathbf{\tilde{\Psi}}^{(k)}$ of $\mathbf{\Psi}^{(k)}$. The squared $\mathcal{L}_2$-loss is then computed and weighted by a learning rate parameter to provide the parameter update $(\delta \bm{w}^{(k)}, \delta \mathbf{b}^{(k)}, \delta \mathbf{c}^{(k)})$. We tried various combinations of training parameters and the best reconstruction results were obtained by applying $15000$ stochastic gradient updates with alternating sampling between the expression classes in the databases. Indeed, we want the network to be able to reconstruct local variations of all possible expressive patterns on an equal foot. We also use a constant learning rate of $0.01$ as well as a weight decay of $0.001$, which seems to provide good results in testing. Finally, we found that adding $25\%$ randomly generated masking noise provided satisfying results. From a manifold learning perspective, the goal of using such denoising criterion is to learn to project corrupted examples (e.g. partially occluded ones, which lie further from the manifold) back on the training data manifold. Such example will be reconstructed closer to the training data and its confidence shall be smaller.

\subsection{Local confidence measurement}\label{confidence}

Given a face image $\mathcal{I}$, we define the point-wise confidence $\alpha^{(k)}(\mathcal{I})$ for point $k$ as the $\mathcal{L}_2$-loss (\textit{i.e.} the reconstruction error) between the HOG descriptor $\mathbf{\Psi}(\mathcal{I})$ extracted from this feature point, and its reconstruction $\mathbf{\tilde{\Psi}}$ outputted by the network, after successively encoding by layers $L_1$ then $L_2$, and decoding in the opposite order. By abuse of notation, we have:

\begin{equation}\label{dec}
\alpha^{(k)}(\mathcal{I}) = ||\mathbf{\Psi}^{(k)}-g^1 \circ g^2 \circ h^2 \circ h^1(\mathbf{\Psi}^{(k)})||^2
\end{equation}

The choice of using an Euclidean distance as a confidence score seems natural as it is optimized during training. We also introduce a confidence measurement $\alpha^{(\tau)}(\mathcal{I})$ defined over triangles $\tau=\{k_1,k_2,k_3\}$ of the facial mesh as:

\begin{equation}\label{dec}
\alpha^{(\tau)}(\mathcal{I}) = \min(\alpha^{(k_1)}(\mathcal{I}),\alpha^{(k_2)}(\mathcal{I}),\alpha^{(k_3)}(\mathcal{I}))
\end{equation}

As highlighted in the following experiments, this triangle-wise confidence measurement can thus be used to weight LEPs to enhance the robustness to partial occlusions to a significant extent.

\section{Local Expression Predictions}\label{local}

\subsection{Learning local trees with random facial masks}\label{learning}

Random Forests (RF) is a popular learning framework introduced in the seminal work of Breiman \cite{breiman2001random}. They have been used to a significant extent in computer vision, and for FER tasks in particular \cite{zhao2014unified,dapogny2015iccv}, due to their ability to nicely handle high-dimensional data such as images as well as being naturally suited for multiclass classification tasks.

In the classical RF framework, each tree of the forest is grown using a subset of training examples (bagging) and a subspace of the input dimension (random subspace). Individual trees are then grown using a greedy procedure that involves, for each node, the generation of a number of binary split candidates that consist in features associated with a threshold. Each candidate thus defines a partition of the labelled training data. The ``best" binary feature is chosen among all features as the one that minimizes an impurity criterion (which is generally defined as either the Shannon entropy or the Gini impurity). Then, the above steps are recursively applied for the left and right subtrees with accordingly rooted data until the label distribution at each node becomes homogeneous, where a leaf node can be set. As stated in \cite{breiman2001random}, the rationale behind training each tree on a random subspace of the input dimension  is that the prediction accuracy of the whole forest depends on both the strength of individual trees and on the independence of the predictions. Thus, by growing individually weaker (e.g. as compared to C4.5) but more decorrelated trees, we can combine these into a more accurate tree collection.

Following this idea, we propose an adaptation of the RF framework that uses spatially-defined Local Subspaces (LS) instead of the traditional Random Subspaces (RS). Each tree is trained using a restricted subspace corresponding to a specific part of the face. The aggregation of local models gives rise to local representations that we call Local Expression Predictions (LEPs). Note that this is not the first time that the output classification predictions of RFs are used as features for a subsequent task. For instance, Ren \textit{et al.} \cite{ren2014face} used local binary features to construct a cascaded feature point alignment method. However, contrary to \cite{ren2014face}, we construct our LEP representation by locally averaging predictions and not by directly using the output prediction of the trees. Furthermore, LEPs offer several advantages over using a set of trees defined on the whole face:

(1) LEPs can be aggregated to provide categorical FER. Those local models (LS-RF) can theoretically capture more diverse information compared to a global one by ``forcing'' the trees to use less informative features, that can still hold some predictive power.

(2) We can use the confidence outputted by the autoencoder network in Section \ref{aec} to weight the LEPs for which the pattern lies further from the training data manifold (WLS-RF). For example, in case of occlusion or drastic illumination changes, we can still use the information from the other face subparts to predict the expression.

(3) LEPs can be used as an intermediate representation for the task of describing Action Units (AUs). Noteworthy, AU classification could benefit from LEPs trained on larger corpus labelled with categorical expressions, as annotation is less time-consuming than FACS coding.

The local trees are trained using Algorithm \ref{lsrfprocedure}. For each tree $t$ in the forest, we generate a face mask $M_t$ defined over triangles $\tau$ on facial feature points of a precomputed mean shape $\bar{f}$. The mask is initialized with a single triangle randomly selected from the mesh. Then, neighbouring triangles are added until the total surface covered by the selected triangles w.r.t. $\bar{f}$ becomes superior to hyperparameter $R$, that represents the (approximate) surface that should be covered by each tree. Finally, tree $t$ is grown on the subspace that corresponds to the facial mask $M_t$.

\begin{algorithm}
\caption{Training Local Subspace Random Forest}
\label{lsrfprocedure}
\textbf{input:} images $\mathcal{I}$ with labels $l$ and feature points $f(\mathcal{I})$
\begin{algorithmic}
\State compute $\bar{f}$, the mean shape
\State compute $s(\tau(\bar{f}))$, surface of triangles $\tau$ on mean shape
\For{$t=1$ to $T$}
\State randomly select a triangle $\tau_i$
\State $r \leftarrow s(\tau_i)$
\State initialize mask $M_t \leftarrow \{\tau_i\}$
\While{$r<R$}
\State draw a list of candidate neighbouring triangles
\State randomly select a triangle $\tau_j$ from that list
\State $r \leftarrow r + s(\tau_j)$
\State $M_t \leftarrow M_t \cup \{\tau_j\}$
\EndWhile
\State randomly select a fraction $\tilde{\mathcal{S}_t} \subset \mathcal{S}$ of subjects
\State balance bootstrap $\tilde{\mathcal{S}_t}$ with downsampling
\State grow tree $t$ on bootstrap $\tilde{\mathcal{S}_t}$ and input subspace $M_t$
\EndFor
\State \textbf{output:} tree predictors $p_t(l | \mathcal{I})$ with associated masks $M_t$
\end{algorithmic}
\end{algorithm}

\subsection{Candidate feature generation}\label{features}

We use a combination of geometric (\textit{i.e.} computed from aligned facial feature points) and appearance features $\phi^{(1)},\phi^{(2)},\phi^{(3)}$, as in \cite{dapogny2015iccv}. Each of these features have different parameters that are generated on-the-fly during training by uniform sampling over their respective variation range.

We use two different geometric feature templates which are generated from the set of $N_p$ facial feature points $f(\mathcal{I})$, provided by an off-the-shelf facial alignment algorithm \cite{xiong2013supervised}. The first of these templates is the Euclidean distance between feature points $f_a$ and $f_b$, normalized w.r.t. intra-ocular distance $iod(f)$ for scale invariance (Equation \ref{pointdist}).

\begin{equation}\label{pointdist}
   \phi^{(1)}_{a,b}(\mathcal{I}) = \frac{||f_a - f_b||_2}{iod(f)}
\end{equation}

Because the feature point orientation is discarded in feature $\phi^{(1)}$ we use the angles between feature points $f_a$, $f_b$ and $f_c$ as our second geometric feature $\phi^{(2)}_{a,b,c,\lambda}$. In order to ensure continuity for angles around $0$, $\phi^{(2)}$ outputs either the cosine or sine of angle $\widehat{f_af_bf_c}$, depending on the value of the boolean parameter $\lambda$ (Equation \eqref{pointangle}):

\begin{equation}\label{pointangle}
   \phi^{(2)}_{a,b,c,\lambda}(\mathcal{I}) = \lambda\cos(\widehat{f_af_bf_c}) + (1-\lambda)\sin(\widehat{f_af_bf_c})
\end{equation}

As appearance features, we use HOGs for their descriptive power and robustness to global illumination changes. Integral HOG feature channels were already computed in Section \ref{networkarchi} for confidence weight computation. Thus, we define feature template $\phi^{(3)}_{\tau,ch,sz,\alpha,\beta,\gamma}$ as an integral histogram computed over channel $ch$ within a window of size $sz$ normalized w.r.t. the intra-ocular distance. Such histogram is evaluated at a point defined by its barycentric coordinates $\alpha$, $\beta$ and $\gamma$ w.r.t. vertices of a triangle $\tau$ defined over feature points $f(\mathcal{I})$.

Each of these candidate features are associated with a set of thresholds $\theta$ to produce binary split candidates. In particular, for each feature template $\phi^{(i)}$, the upper and lower bounds are estimated from training data beforehand and candidate thresholds are drawn from a uniform distribution in the range of these values.

\subsection{Occlusion-robust expression recognition}\label{localtest}

When testing, a face image $\mathcal{I}$ is successively rooted left or right for each tree $t$ depending of the outputs of the binary tests stored in the tree nodes, until it reaches a leaf. The tree $t$ thus outputs a probability vector $p_t(l | \mathcal{I})$ whose components are either $1$ for the represented class, or $0$ otherwise. Prediction probabilities are then averaged among the $T$ trees of the forest (Equation \eqref{probasRSRF}).

\begin{equation}\label{probasRSRF}
   p(l | \mathcal{I}) = \frac{1}{T}\sum \limits_{t=1}^{T}{p_t(l | \mathcal{I})}
\end{equation}

Those prediction probabilities are computed similarly for the global RF (RS-RF) and the LS-RF. However, for LS-RF the output probabilities of the trees have some degrees of locality and we can write the above formula as a sum over local probabilities defined for each triangle (Equation \eqref{probasLSRF}).

\begin{equation}\label{probasLSRF}
   p(l | \mathcal{I}) = \frac{1}{T}\sum \limits_{\tau} {Z_\tau} p(l | \mathcal{I},\tau)
\end{equation}

Where $p(l | \mathcal{I},\tau)$ is the Local Expression Prediction (LEP) probability vector associated with triangle $\tau$ on the facial mesh:

\begin{equation}\label{pau}
p(l | \mathcal{I},\tau) = \frac{1}{Z_\tau}\sum \limits_{t=1}^{T}\frac{\delta(\tau \in M_t)p_t(l | \mathcal{I})}{|M_t|}
\end{equation}

With $\delta(\tau \in M_t)$ being a function that returns $1$ if triangle $\tau$ belongs to mask $M_t$, and $0$ otherwise. $|M_t|$ is the number of times tree $t$ is used in Equation \eqref{probasLSRF}, and $Z_\tau$ is the sum of prediction values for all expression classes $l$. Thus, a global expression probability is defined by a (normalized) sum of LEPs. Note that those LEP vectors $p(l | \mathcal{I},\tau)$ are not strictly limited to triangle $\tau$ but defined within its neighbourhood, with a radius that depends on hyperparameter $R$. The setting of $R$ thus controls the locality of the trees, as it will be discussed in the experiments.

Moreover, LEPs can be weighted by local confidence measurements to give rise to the Weighted Local Subspace Random Forest model (WLS-RF): 

\begin{equation}\label{probasWLSRF}
   p(l | \mathcal{I}) = \frac{\sum \limits_{\tau} \alpha^{(\tau)} Z_\tau p(l | \mathcal{I},\tau)}{\sum \limits_{\tau} \alpha^{(\tau)} Z_\tau}
\end{equation}

Where $\alpha^{(\tau)}$ is the triangle-wise confidence measurement that is outputted by the autoencoder network described in Section \ref{aec}, for triangle $\tau$. This weighting scheme allows to better handle partial occlusions, by downweighting the local RFs associated with the most unreliable appearance patterns.

\subsection{Action Unit detection}\label{pautoau}

\subsubsection{From LEPs to Action Units} 
\begin{figure}[htbf]
\centering
\includegraphics[width=\linewidth]{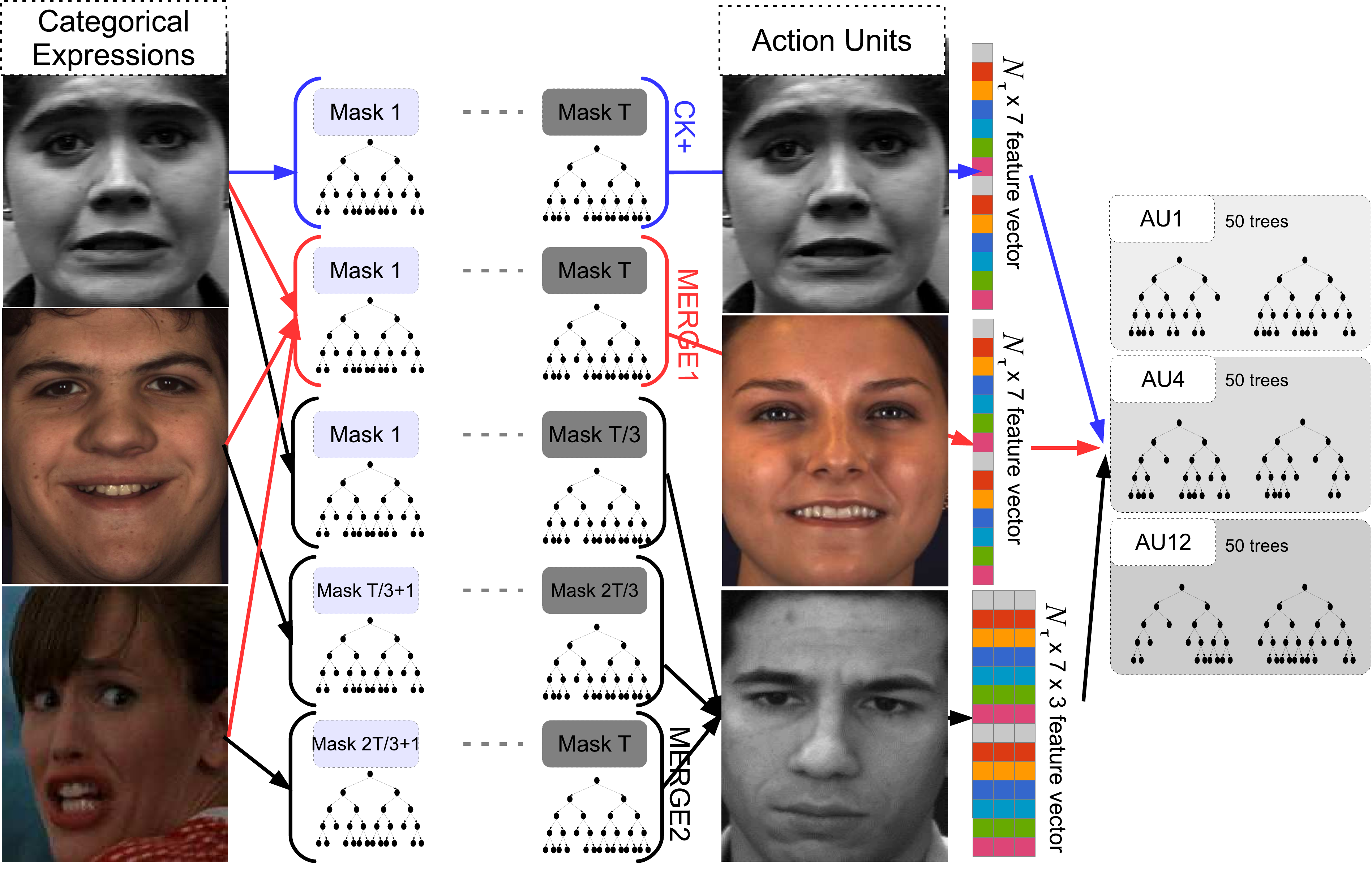}
\caption{AU recognition using LEP features.}
\label{pautoaufig}
\end{figure}

LEPs are local responses related to categorical facial expressions. Thus, it makes sense to assume that LEPs can somehow be related to AUs and constitute a good high-level representation for AU recognition. To this end, Figure \ref{pautoaufig} describes the AU recognition framework, in which LEP vectors corresponding to each triangle are extracted by a first layer of local trees, trained on a categorical expression dataset. The concatenation of all LEP vectors $p(l | \mathcal{I},\tau)$ for every expression $l$ (6 universal expressions plus the neutral one) and triangle $\tau$ of the facial mesh gives rise to a $7 \times N_{\tau}$ feature vector used by a second layer of trees defined for each AU (with $N_{\tau}$ the number of triangles of the facial mesh).  Thus, the AU recognition layer is trained on a FACS-labelled dataset using only one feature template $\phi^{(0)}_{l,\tau} = p(l | \mathcal{I},\tau)$, with associated thresholds $\theta$ randomly generated from a uniform distribution in the $[0;1]$ interval.

As illustrated on Figure \ref{pautoaufig}, we also study the importance of using multiple available expression datasets for learning the first layer of trees (\textit{i.e.} LEP representation). We can either train the models on a specific categorical expression database, or merge the datasets to learn LEP representation from all the available corpus ($M1$). Finally, we can also learn LEPs separately from the different categorical expression datasets and use a concatenation of the LEP feature vectors as an input for the second (AU prediction) tree layer ($M2$). Section \ref{AUsexp} shows that those two approaches enhance the predictive power of the AU detection framework. Furthermore, those two strategies can complement each other well. Indeed, $M1$ requires to simultaneously load multiple datasets at training time, $M2$ involves computing multiple LEP features for evaluation. Thus, a combination of those two strategies can be used to fulfil the target memory/time requirements.

Also note that we voluntarily keep the AU recognition layer simple so as to showcase the usefulness of LEP representation for the AU prediction task, as compared to low-level engineered descriptors and other state-of-the-art methods. However, as shown in other works on expression recognition \cite{nicolle2015facial}, recent approaches such as multi-task formulations (e.g. training a single RF for predicting multiple AUs) can significantly improve performances.

\subsubsection{Confidence assessment for AU prediction} 
\begin{figure*}[htbf]
\centering
\includegraphics[width=\linewidth]{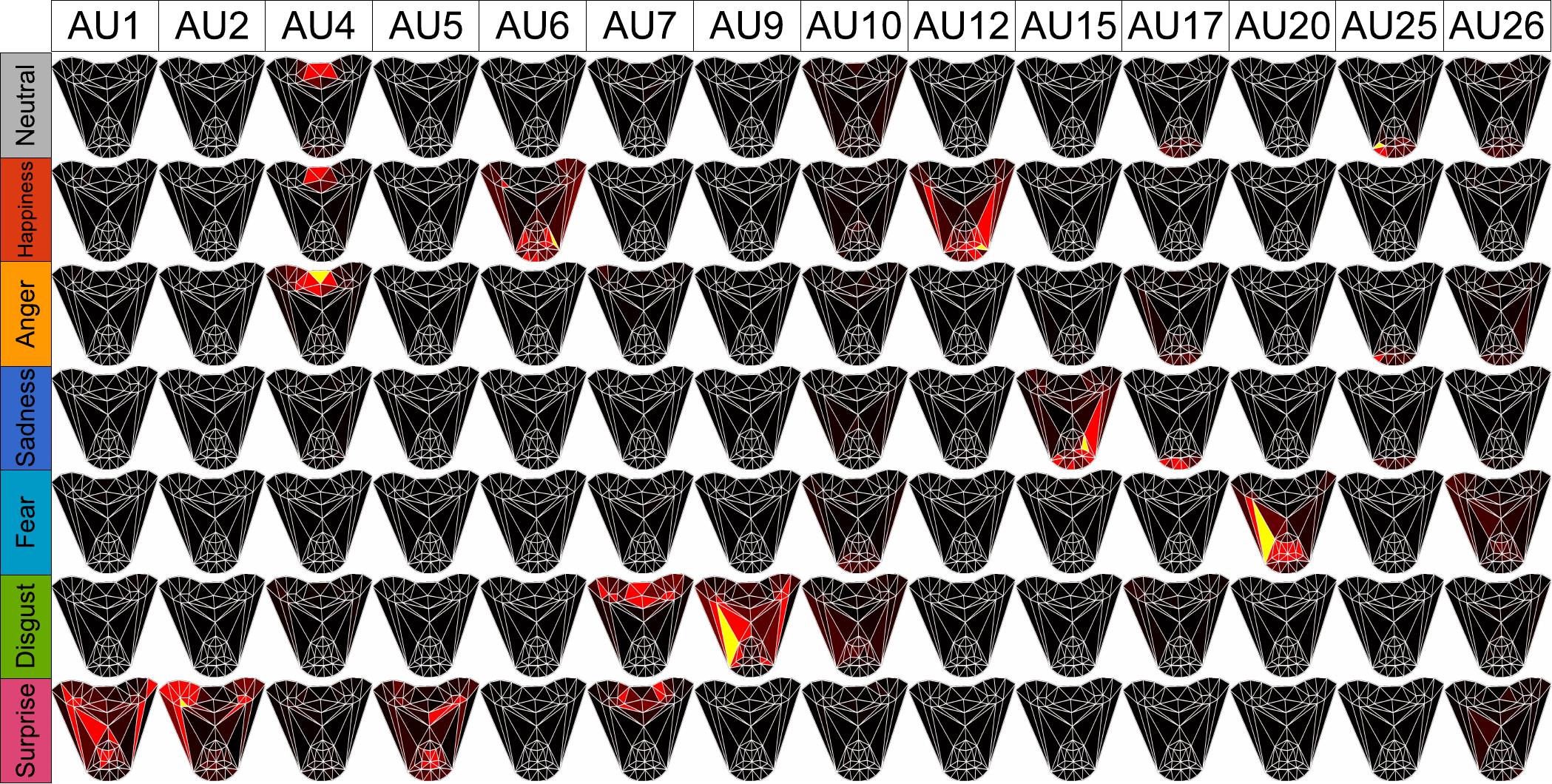}
\caption{LEPs heat map for each of the 14 AUs (CK+ database). Only top-level LEP features are displayed for each AU. Best viewed in color.}
\label{heatmap}
\end{figure*}

Because AUs are defined locally, chances are that AU activation relatively to an occluded area can not be predicted at all. Thus, we use the weights outputted by the autoencoder network to automatically derive a confidence score relatively to each AU indexed by $m$. To this end, we define as $N^{(m)}_{l,\tau}$ the number of times that the LEP feature $\phi^{(0)}_{l,\tau}$ was selected for splitting at the root of the trees, among all trees in the forest. This is highlighted on Figure \ref{heatmap}. The reason for exclusively considering features at the root of the trees is that those features are selected from large numbers of training examples, as opposed to features from nodes deeper in the trees, that are essentially more noisy.

Note that, while most approaches focus on describing expressions as a combination of AUs, we can decompose each AU as a set of local expression predictions. For example, for AU1 (inner brow raiser) and AU2 (outer brow raiser), the most relevant LEPs are triangles corresponding to the inner and outer brows, associated with expression \textit{surprise}, respectively. AU4 (brow lowerer) mainly uses triangles between the eyes associated with expression \textit{anger}. AU9 (nose wrinkler) mainly uses triangles from the nose and cheek regions, associated with \textit{disgust}. AU12 (lip corner puller) and AU20 (lip stretcher) respectively use triangles corresponding to lip corners with expressions \textit{happiness} and \textit{fear}.

We then define the AU-specific confidence measurement $\alpha_m$ for AU $m$ as the sum of confidences $\alpha^{(\tau)}$ of triangles $\tau$ of the facial mesh, weighted by the proportion of LEP features from that triangle, that are used to describe the activation of AU $m$:

\begin{equation}\label{probasWLSRF}
   \alpha_m = \frac{\sum_{\tau}\alpha^{(\tau)}{N^{(m)}_{l,\tau}}}{\sum_{\tau}N^{(m)}_{l,\tau}} 
\end{equation}

Thus, the AU-specific confidence measurement is proportional to the confidence of the face regions that are the most useful for describing the activation of a specific AU. We show in the following section that such simple setting allows to highlight the cases were the AU predictions are deemed unreliable.

\section{Experiments}\label{exper}

In this section, we evaluate our approach on several FER benchmarks. Section \ref{datas} introduces the databases that are used for test. Section \ref{expset} sums up our experimental protocols and describe hyperparameter settings to ensure reproducibility of the results. In Section \ref{clean}, we show results for FER on non-occluded data on three publicly available FER benchmarks that exhibit various degrees of difficulty, showing that our approach improves the state-of-the-art. Then, in Section \ref{corr}, we report results on synthetically occluded images. Thus, we can precisely measure the robustness of our approach to occlusions. In Section \ref{AUsexp} we give results of AU detection, showing that LEPs yield high predictive power for the task of AU detection compared to low-level features or state-of-the-art approaches. We also evaluate the relevance of the AU-specific confidence measurements. Lastly, Section \ref{realtime} reports evidence of the real-time capacities of the proposed framework.

\subsection{Datasets}\label{datas}

\subsubsection{Categorical expressions datasets}\label{catdata}

\textbf{The CK+} or \textbf{Extended Cohn-Kanade database \cite{lucey2010extended}} contains 123 subjects, each one displaying some of the 6 universal expressions (\textit{anger}, \textit{happiness}, \textit{sadness}, \textit{fear}, \textit{digust} and \textit{surprise}) plus the non-basic expression \textit{contempt}. Expressions are prototypical and performed in a controlled lab environment with no head pose variation. As it is done in other approaches, we use the first (\textit{neutral}) and three apex frames for each of the 327 sequences for 8-class categorical FER. As some approaches discard the frames labelled as \textit{contempt}, we also report 7-class accuracy from 309 sequences.

\textbf{The BU-4D} or \textbf{BU-4DFE database \cite{yin2008high}} contains 101 subjects, each one displaying 6 acted categorical facial expressions with moderate head pose variations. Expressions are still prototypical but they are performed with lower intensity and greater variability than in CK+, hence the lower baseline accuracy. Sequence duration is about 100 frames. As the database does not contain frame-wise expression, we manually select neutral and apex frames for each sequence.

\textbf{The SFEW} or \textbf{Static Facial Expression in the Wild database \cite{dhall2011static}} contains $700$ images from $95$ subjects displaying $7$ facial expressions in a real-world environment. Data was gathered from video clips using a semi-automatic labelling process. The strictly person-independent evaluation (SPI) benchmark is composed of two folds of (roughly) same size. As done in other approaches, we report cross-validation results averaged over the two folds.

\subsubsection{Action Unit datasets}\label{AUdata}

\textbf{The CK+ database} is also FACS-annotated, therefore we report results for the recognition of 14 of the most common AUs (AU1,2,4,5,6,7,9,10,12,15,17,20,25,26).

\textbf{The BP4D database \cite{zhang2014bp4d}} contains 41 subjects. Each subject was asked to perform 8 tasks, each one supposed to give rise to 8 spontaneous expressions (\textit{anger}, \textit{happiness}, \textit{sadness}, \textit{fear}, \textit{digust}, \textit{surprise}, \textit{embarrassment} or \textit{pain}). In \cite{zhang2014bp4d} the authors extracted subsequences of about 20 seconds for manual FACS annotations, arguing that these subsets contain the most expressive behaviors. As done in the litterature \cite{zhang2014bp4d} we report results for recognition of 12 AUs (1,2,4,6,7,10, 12,14,15,17,23,24). We randomly extract $10000$ images for training and evaluate the AU classifiers on the whole dataset.

\textbf{The DISFA} or \textbf{Denver Intensity of Spontaneous Facial Actions \cite{mavadati2013disfa}} contains videos of 27 subjects with different ethnicities and genders that were recorded watching a 4-minute emotive video stimulus. Data have been manually labeled frame by frame for 12 AUs (1,2,4,5,6,9,12,15,17,20, 25,26) on a 6-level scale by a human expert, and verified by a second FACS coder. For the purpose of predicting AU occurrence, we consider AU which intensity is below 1 as non-activated. We randomly extract $6292$ images for training and test on the $125832$ images.

\subsection{Experimental setup}\label{expset}

\subsubsection{Evaluation metrics} For both occluded and non-occluded scenarios of categorical FER we use the overall accuracy as a performance metric. We also report confusion matrices to show the discrepancies between recognition of the expression classes. For AU detection we use the area under the ROC curve (AUC) as a performance metric, as it is widely used in the literature because it is independent of the setting of a decision threshold. For all the experiments, RF classifiers are evaluated with Out-Of-Bag (OOB) error estimate, with bootstraps generated at the subject level to ensure that, for each tree, subjects used for training are not used for testing this specific tree. The OOB error, according to \cite{breiman2001random}, is an unbiased estimate of the true generalization error. Moreover, as stated in \cite{bylander2002estimating} this estimate is generally more pessimistic than traditional (e.g. k-fold or leave-one-subject-out) cross-validation estimates, further reflecting the quality of the results. For AU recognition, LEPs are generated for Out-Of-Bag examples for each tree and AUs are evaluated with OOB error. 


\subsubsection{Hyperparameter setting} In order to decrease the variance of the error we train large collections of trees ($T_1=1000$ for LEP generation, $T_2=50$ for AU detection). For training the local models, we set the locality parameter $R$ to 0.1 (which means that each local model uses $1/10$ of the face total surface) which provided good robustness to occlusions. Finally, we use $40$ $\phi^{(1)}$, $40$ $\phi^{(2)}$ and $160$ $\phi^{(3)}$ features for learning LEPs, as well as $25$ threshold evaluations per features. For AU detection, we examine 100 $\phi^{(0)}$ features at each node, each associated with $25$ threshold values. Note however that the values of these hyperparameters (except for $R$) had very little influence on the performances. This is due to the complexity of the RF framework, in which individually weak trees (e.g. that are grown by only examining a few features per node) are generally less correlated, still outputting decent predictions when combined altogether. 

For the occluded scenarios on CK+ and BU4D, the autoencoder networks are trained in a cross-database fashion (\textit{i.e.} training on CK+ and testing on BU4D and vice versa). Lastly, on SFEW database, we use the autoencoder network trained on CK+, as SFEW embraces multiple examples of occluded faces.

\subsection{Categorical FER}\label{clean}

\subsubsection{FER on non-occluded images}\label{clean}

In Tables \ref{compck}, \ref{compbu}, \ref{compsfew} we report the average accuracy obtained by our local subspace Random Forest (LS-RF) and the confidence-weighted version (WLS-RF). We also compare with standard RF (RS-RF).

\begin{table*}[t]
\centering
\begin{minipage}{0.3\textwidth}
\vspace{0pt}
\centering
\caption{CK+ database. $^{\dagger}$: CK database}
\scalebox{0.75}{
\begin{tabular}{ l | c | c | r }
	CK+ & Protocol & 7em & 8em\\
	\hline
	LBP \cite{shan2009facial} & 10-fold & $88.9 ^{\dagger}$ & - \\
	\hline
	CSPL \cite{zhong2012learning} & 10-fold & $89.9 ^{\dagger}$ & - \\
	\hline
	iMORF \cite{zhao2014unified}& 10-fold & - & 90.0 \\
	\hline
	AUDN \cite{liu2015inspired}& 10-fold & 93.7 & 92.0 \\
  \hline
	RS-RF & OOB & 92.6 & 91.5 \\
  \hline
	LS-RF & OOB & 94.1 & \textbf{93.4}\\
  \hline
	WLS-RF & OOB & \textbf{94.3} & \textbf{93.4}\\
	\hline
\end{tabular}}
\label{compck}
\end{minipage}\hfill
\begin{minipage}{0.35\textwidth}
\vspace{0pt}
\centering
\caption{BU4D database}
\scalebox{0.85}{
\begin{tabular}{ l | c | r }
	BU4D & Protocol & \% Acc\\
	\hline
	BoMW \cite{xu2010automatic}& 10-fold & $63.8$ \\
	\hline
	Geometric \cite{sun2008facial}& 10-fold & $68.3$ \\
	\hline
	LBP-TOP \cite{hayat2012evaluation}& 10-fold & 71.6 \\
	\hline
	2D FFDs \cite{sandbach2011dynamic}& 10-fold & 73.4 \\
  \hline
	RS-RF & OOB & 73.1 \\
  \hline
	LS-RF & OOB & 74.3\\
  \hline
	WLS-RF & OOB & \textbf{75.0}\\
	\hline
\end{tabular}}
\label{compbu}
\end{minipage}\hfill
\begin{minipage}{0.29\textwidth}
\vspace{0pt}
\centering
\caption{SFEW database}
\scalebox{0.85}{
\begin{tabular}{ l | r }
	SFEW & \% Acc\\
	\hline
	PHOG-LPQ \cite{dhall2011static} & 19.0 \\
	\hline
	DS-GPLVM \cite{eleftheriadis2015discriminative} & 24.7 \\
	\hline
	AUDN \cite{liu2015inspired} & 30.1 \\
  \hline
	Semi-Supervised \cite{liu2013enhancing} & 34.9 \\
  \hline
	RS-RF & 35.7\\
  \hline
	LS-RF & 35.6\\
  \hline
	WLS-RF & \textbf{37.1}\\
	\hline
\end{tabular}}
\label{compsfew}
\end{minipage}\hfill
\end{table*}

\begin{table*}[t]
\centering
\begin{minipage}{0.35\textwidth}
\vspace{0pt}
\centering
\caption{Confusion matrix (CK+-8em)}
\scalebox{0.45}{
\begin{tabular}{c | c c c c c c c c}
	\hline
	\rowcolor{Gray1}
	& ne & ha & an & sa & fe & di & co & su\\
	\hline
	\rowcolor{Blue5}
	\cellcolor{Gray1}ne & \cellcolor{Blue1} 92.4 & \cellcolor{Blue45}0.3 & \cellcolor{Blue45}0.9 & \cellcolor{Blue5}0.6 & \cellcolor{Blue45}1.2 & \cellcolor{Blue45}0.6 & \cellcolor{Blue4}3.97 & 0\\
	\rowcolor{Blue5}
	\cellcolor{Gray1}ha & 0 & \cellcolor{Blue1}100 & 0 & 0 & 0 & 0 & 0 & 0\\
	\rowcolor{Blue5}
	\cellcolor{Gray1}an & \cellcolor{Blue4}4.4 & 0 & \cellcolor{Blue1}91.1 & 0 & 0 & \cellcolor{Blue45}2.3 & \cellcolor{Blue45}2.3 & 0\\
	\rowcolor{Blue5}
	\cellcolor{Gray1}sa & \cellcolor{Blue34}22.6 & 0 & 0 & \cellcolor{Blue3} 77.4 & 0 & 0 & 0 & 0\\
	\rowcolor{Blue5}
	\cellcolor{Gray1}fe & \cellcolor{Blue45}1.3 & \cellcolor{Blue4}4 & 0 & 0 & \cellcolor{Blue1}90.7 & 0 & 0 & \cellcolor{Blue4}4\\
	\rowcolor{Blue5}
	\cellcolor{Gray1}di & \cellcolor{Blue4}3.4 & 0 & \cellcolor{Blue45}0.6 & 0 & 0 & \cellcolor{Blue1}96.1 & 0 & 0\\
	\rowcolor{Blue5}
	\cellcolor{Gray1}co & \cellcolor{Blue34}11.1 & 0 & 0 & \cellcolor{Blue4}3.7 & 0 & 0 & \cellcolor{Blue2}85.2 & 0\\
	\rowcolor{Blue5}
	\cellcolor{Gray1}su & \cellcolor{Blue45}1.6 & 0 & 0 & 0 & \cellcolor{Blue45}0.4 & 0 & \cellcolor{Blue45}1.2 & \cellcolor{Blue1}96.8\\
	
	\hline
\end{tabular}}
\label{confck}
\end{minipage}\hfill
\begin{minipage}{0.32\textwidth}
\vspace{0pt}
\centering
\caption{Confusion matrix (BU4D)}
\scalebox{0.5}{
\begin{tabular}{c | c c c c c c c}
	\hline
	\rowcolor{Gray1}
	& ne & ha & an & sa & fe & di & su\\
	\hline
	\rowcolor{Blue5}
	\cellcolor{Gray1}ne & \cellcolor{Blue1} 89.5 & 0 & \cellcolor{Blue5}1.8 & \cellcolor{Blue45}4.4 & \cellcolor{Blue5}0.9 & \cellcolor{Blue5}0.9 & \cellcolor{Blue45}2.6 \\
	\rowcolor{Blue5}
	\cellcolor{Gray1}ha & \cellcolor{Blue5}2 & \cellcolor{Blue1}89.9 & 0 & 0 & \cellcolor{Blue45} 5 & \cellcolor{Blue5}2 & \cellcolor{Blue5}1\\
	\rowcolor{Blue5}
	\cellcolor{Gray1}an & \cellcolor{Blue4}10.1 & 0 & \cellcolor{Blue2}70.7 & \cellcolor{Blue4}7.1 & \cellcolor{Blue5}2 & \cellcolor{Blue4}9.1 & \cellcolor{Blue5}1\\
	\rowcolor{Blue5}
	\cellcolor{Gray1}sa & \cellcolor{Blue4}11 & 0 & \cellcolor{Blue34}15 & \cellcolor{Blue2} 71 & \cellcolor{Blue5}3 & 0 & 0\\
	\rowcolor{Blue5}
	\cellcolor{Gray1}fe & \cellcolor{Blue4}9.8 & \cellcolor{Blue34}17.6 & \cellcolor{Blue5}2.9 & \cellcolor{Blue45}5.9 & \cellcolor{Blue3}38.3 & \cellcolor{Blue4}11.8 & \cellcolor{Blue4}13.7\\
	\rowcolor{Blue5}
	\cellcolor{Gray1}di & \cellcolor{Blue5}3 & \cellcolor{Blue45}4 & \cellcolor{Blue45}6.9 & \cellcolor{Blue5}1 & \cellcolor{Blue45}7.9 & \cellcolor{Blue2}73.3 & \cellcolor{Blue45}4\\
	\rowcolor{Blue5}
	\cellcolor{Gray1}su & 0 & \cellcolor{Blue5}1 & 0 & \cellcolor{Blue5}1 & \cellcolor{Blue45}6.2 & 0 & \cellcolor{Blue1}91.8\\
	\hline
\end{tabular}}
\label{confbu}
\end{minipage}\hfill
\begin{minipage}{0.32\textwidth}
\vspace{0pt}
\centering
\caption{Confusion matrix (SFEW)}
\scalebox{0.5}{
\begin{tabular}{c | c c c c c c c}
	\hline
	\rowcolor{Gray1}
	& ne & ha & an & sa & fe & di & su\\
	\hline
	\rowcolor{Blue5}
	\cellcolor{Gray1}ne & \cellcolor{Blue3} 50.2 & \cellcolor{Blue45}8.8 & \cellcolor{Blue45}9.0 & \cellcolor{Blue45}10.0 & \cellcolor{Blue5}2.0 & \cellcolor{Blue4}16.9 & \cellcolor{Blue5}3.1 \\
	\rowcolor{Blue5}
	\cellcolor{Gray1}ha & \cellcolor{Blue45}10.6 & \cellcolor{Blue23}67.5 & \cellcolor{Blue45}6.2 & \cellcolor{Blue45}6.9 & \cellcolor{Blue45}2.6 & \cellcolor{Blue45}3.5 & \cellcolor{Blue45}2.6\\
	\rowcolor{Blue5}
	\cellcolor{Gray1}an & \cellcolor{Blue4}25.4 & \cellcolor{Blue4}16.1 & \cellcolor{Blue34}31.3 & \cellcolor{Blue45}10.1 & \cellcolor{Blue45}3.7 & \cellcolor{Blue5}0.9 & \cellcolor{Blue45}12.5\\
	\rowcolor{Blue5}
	\cellcolor{Gray1}sa & \cellcolor{Blue4}21.2 & \cellcolor{Blue4}21.2 & \cellcolor{Blue45}8.1 & \cellcolor{Blue4}22.2 & \cellcolor{Blue45}7.1 & \cellcolor{Blue45}9.1 & \cellcolor{Blue45}11.1\\
	\rowcolor{Blue5}
	\cellcolor{Gray1}fe & \cellcolor{Blue45}14.2 & \cellcolor{Blue4}16.2 & \cellcolor{Blue45}13.0 & \cellcolor{Blue45}5.0 & \cellcolor{Blue4}23.1 & \cellcolor{Blue45}7.1 & \cellcolor{Blue4}21.3\\
	\rowcolor{Blue5}
	\cellcolor{Gray1}di & \cellcolor{Blue34}31.3 & \cellcolor{Blue4}23.7 & \cellcolor{Blue45}10.4 & \cellcolor{Blue45}7.1 & \cellcolor{Blue45}3.7 & \cellcolor{Blue4}15.6 & \cellcolor{Blue45}8.2\\
	\rowcolor{Blue5}
	\cellcolor{Gray1}su & \cellcolor{Blue45}15.4 & \cellcolor{Blue45}11.0 & \cellcolor{Blue45}12.1 & \cellcolor{Blue5}3.3 & \cellcolor{Blue45}7.7 & \cellcolor{Blue45}6.6 & \cellcolor{Blue3}44.0\\
	\hline
\end{tabular}}
\label{confsfew}
\end{minipage}\hfill
\end{table*}

\begin{figure}[htbf]
\centering
\includegraphics[width=0.65\linewidth]{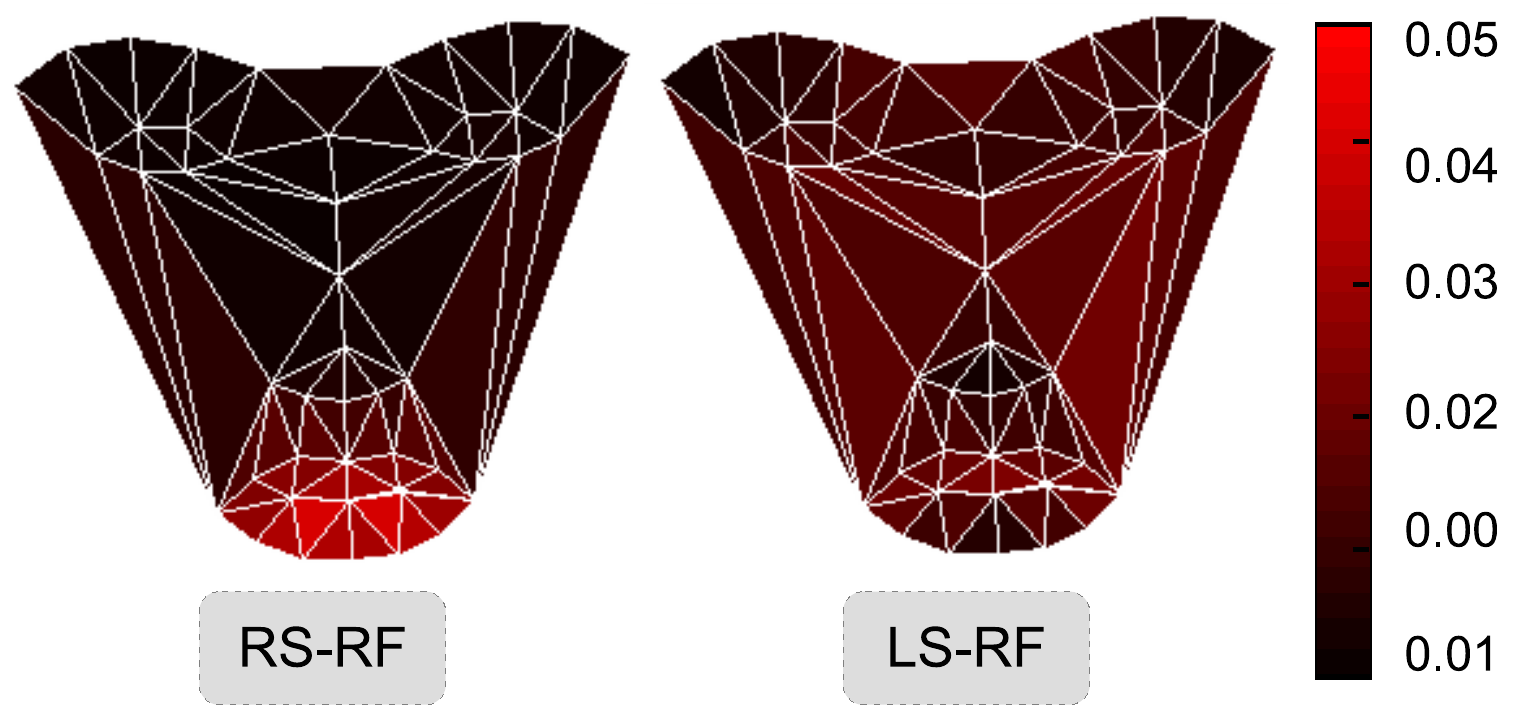}
\caption{Proportion of top-level (tree root) features per triangle. Best viewed in color.}
\label{featurerep}
\end{figure}

Generally speaking, classification results of LS-RF are a little better than those of the RS-RF. Indeed, forcing the trees to be local allows to capture more diverse information. RS-RF relies quite heavily on the mouth region, but other areas (e.g. around the eyes, eyebrows and nose regions) may also convey information that can be captured by local models. Figure \ref{featurerep} displays the proportion of top-level features over all triangles of the face area.

While more than $90\%$ of the features extracted by RS-RF are concentrated around the mouth, the repartition for LS-RF is more homogeneous. Hence, LS-RF is less prone to a misalignment of the mouth feature points, or to occlusions of the mouth region. Furthermore, weighting the local predictions (WLS-RF) using the confidence score from the autoencoder network allows to enhance the results on BU4D and SFEW. The reason is that subjects from those datasets exhibit uncommon morphological traits, occlusion or lighting patterns. As such, more emphasis is put on reliable local patterns, resulting in a better overall accuracy. It also explains why the accuracy is equivalent for LS-RF and WLS-RF on CK+ database, where there is less variability. On the three databases, LS-RF and WLS-RF models provide better results compared to state-of-the-art approaches, even though some of these use complex FFD or spatio-temporal features (LBP-TOP), or use additional unlabelled data for regularization \cite{liu2013enhancing}. Note however that the evaluation protocols are different for some of these approaches. For example, authors in \cite{eleftheriadis2015discriminative} use only the texture information and not the provided landmarks.

Tables \ref{confck}, \ref{confbu}, \ref{confsfew} show the confusion matrices of WLS-RF on CK+, BU4D and SFEW respectively. Generally speaking, expressions \textit{neutral}, \textit{happy} and \textit{surprise} are mostly correctly recognized, as they involve the most recognizable patterns (smile or eyebrow raise). \textit{Anger} and \textit{disgust} are also accurately recognized on CK+ and BU4D but not so much on SFEW. \textit{Sadness} and \textit{fear} seems to be the most subtle ones, particularly on BU4D and SFEW where those expressions can be misclassified as \textit{surprise} or \textit{happy}, respectively.

\subsubsection{FER on occluded face images}\label{corr}

\begin{figure}[htbf]
\centering
\includegraphics[width=0.65\linewidth]{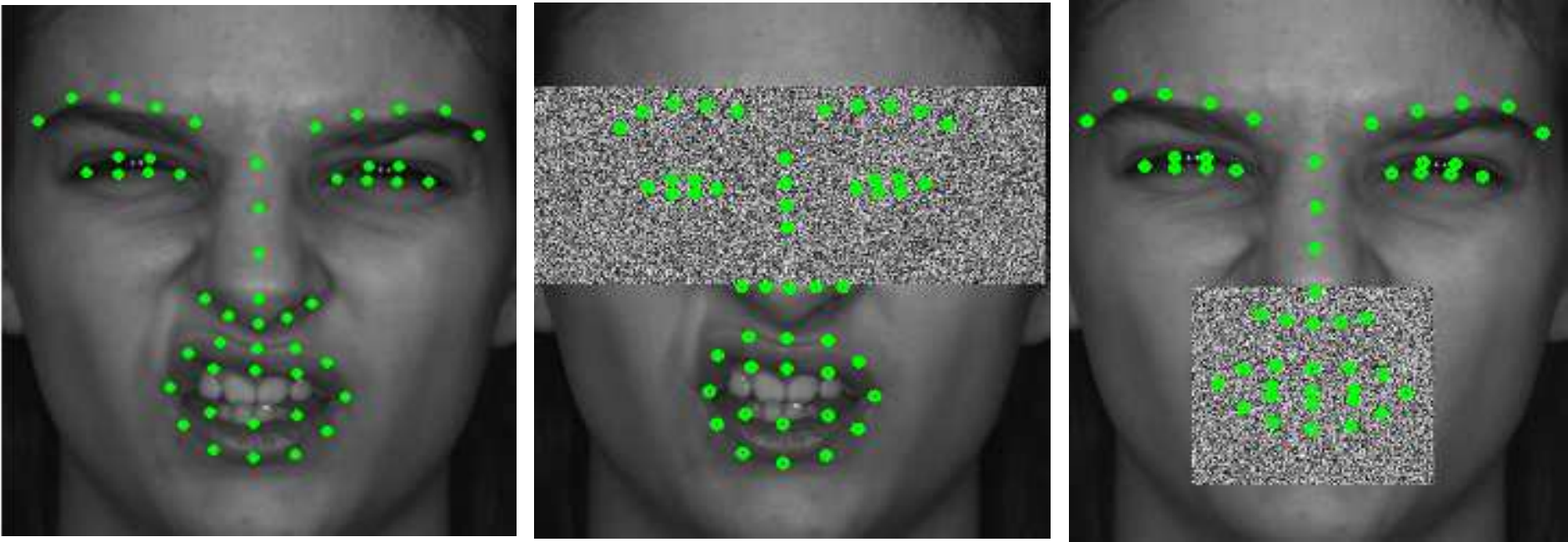}
\caption{Examples of occluded faces from BU4D with aligned feature points. Left: non-occluded, middle: eyes occluded, right: mouth occluded. Also notice how the presence of an occlusion may have a critical effect on the quality of the feature point alignment.}
\label{occex}
\end{figure}

In order to assess the robustness of our system to partial face occlusion, we measured the average accuracy outputted by RS-RF, LS-RF and WLS-RF on CK+ (8 expressions) and BU4D (7 expressions) databases with synthetic occlusions. More precisely, for each image we use the feature points tracked on non-occluded images to highlight the eyes and mouth regions. We then overlay a noisy pattern (see Figure \ref{occex}), which is a more challenging setup than black boxes used in \cite{zhang2014random,huang2012towards}. We add margins of $20$ pixels to the bounding boxes to make sure we cover the whole eyes (with eyebrows, as it represents the most valuable source of information from the eye region) and mouth region. Finally, we align the feature points on the occluded sequences.

\begin{figure*}[htbf]
\centering
\includegraphics[width=\linewidth]{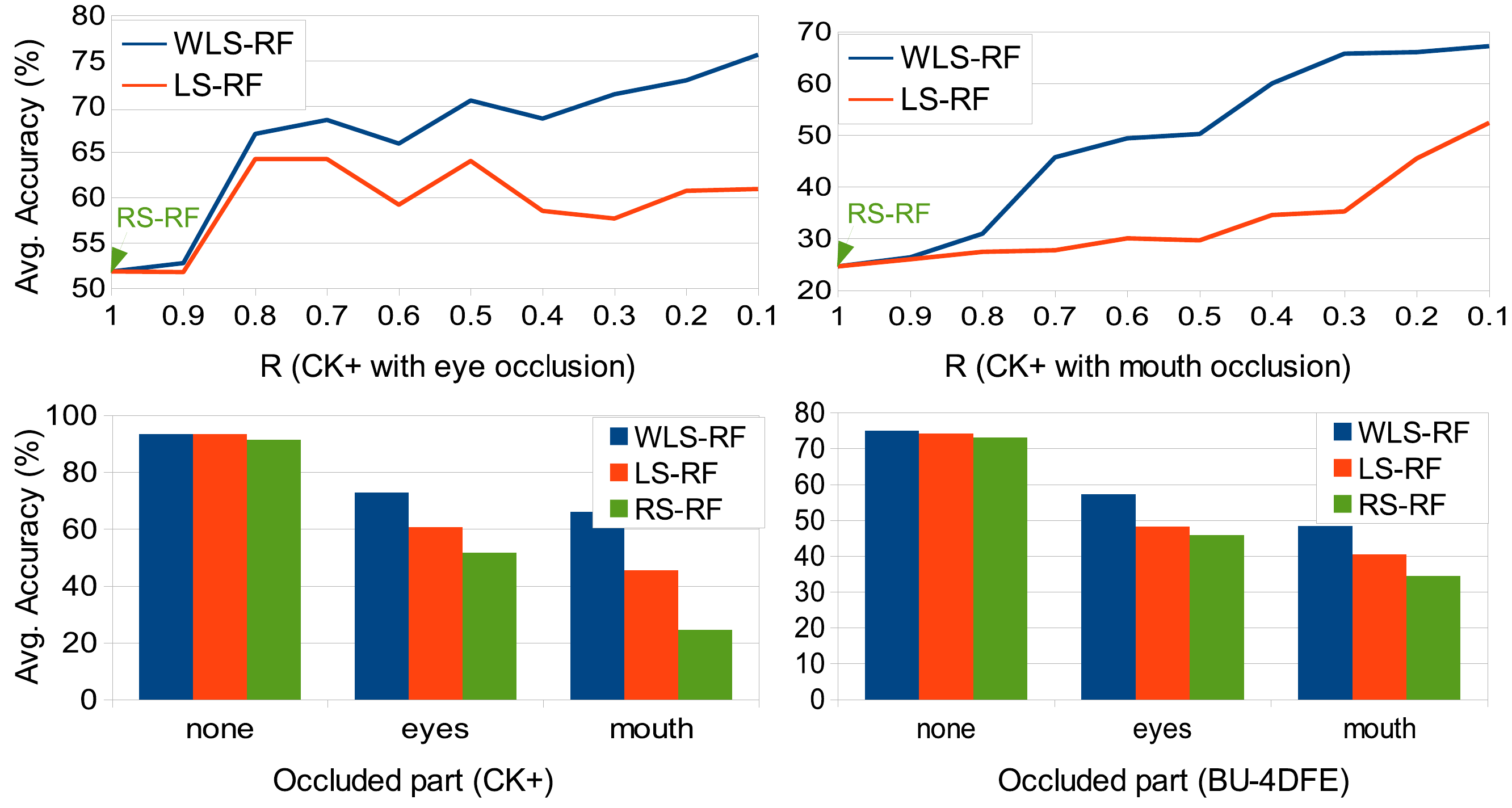}
\caption{Accuracy outputted on occluded CK+ and BU4D databases}
\label{occlres}
\end{figure*}


\paragraph{Influence of $R$:} Graphs of Figure \ref{occlres} show the variation of average accuracy \textit{vs.} hyperparameter $R$ that controls the locality of the trees, respectively under eyes and mouth occlusion on CK+ database. Performances of RS-RF fall heavily when the mouth is occluded (from $91.5\%$ to $25.4\%$), as observed in \cite{zhang2014random}. This further proves that the global model relies essentially on mouth features to decipher facial expressions. Forcing the trees to be more local (e.g. setting $R$ to $0.1$ or $0.2$) allows to capture more diverse cues from multiple facial areas, ensuring more robustness to mouth occlusion. It also explains why LS-RF models with $R=0.8-0.5$ can already be quite robust to eyes occlusions, as the majority of the information used on such models likely comes from mouth area. Nevertheless, on those two occlusion scenarios, WLS-RF achieves a substantially better accuracy than the unweighted local models. Figure \ref{occlres} also shows the accuracy comparison for both eyes and mouth occlusion scenarios on CK+ and BU4D, with $R=0.1$. On the two databases, LS-RF is more robust to partial occlusions than RS-RF. Furthermore, WLS-RF also provides better accuracy than both LS-RF and RS-RF. Overall, the recognition percentage for WLS-RF under mouth occlusion is $67.1\%$ against $30.3\%$ for \cite{zhang2014random} in the case where classifiers are trained on non-occluded faces and tested on occluded ones.

\begin{figure*}[htbf]
\centering
\includegraphics[width=\linewidth]{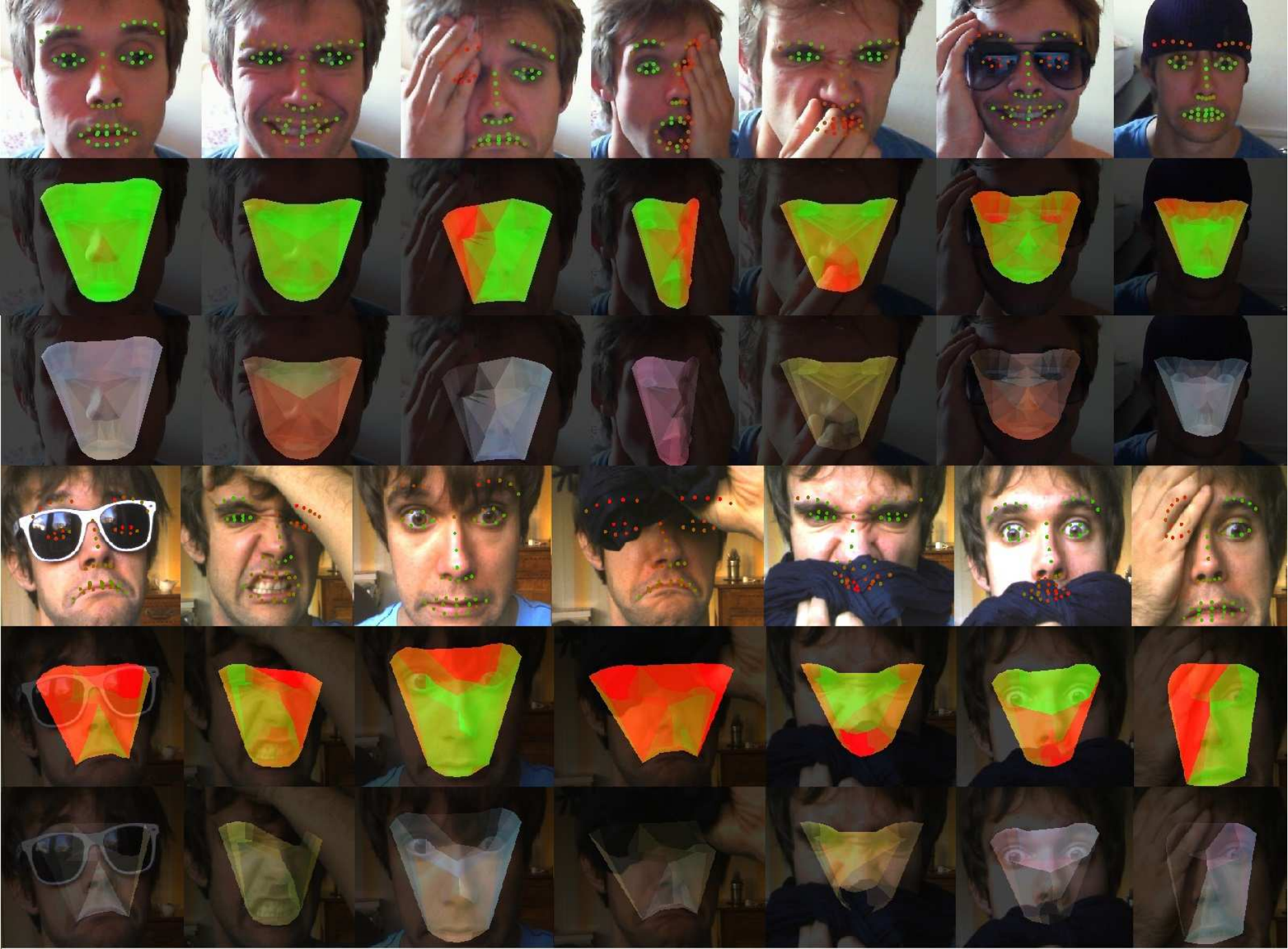}
\caption{Examples of local FER under realistic occlusions. Top rows: point-wise confidence scores (red: low confidence, green: high confidence). Middle rows: triangle-wise scores. Bottom rows: weighted local classification (transparent for low confidences, gray for neutral, red for \textit{happy}, yellow for \textit{angry}, blue for \textit{sad}, cyan for \textit{fear}, green for \textit{disgust} and magenta for \textit{surprise}). Best viewed in color.}
\label{qualitative}
\end{figure*}

\paragraph{Realistic occlusions:} our occlusion model is however quite ``boring'', in the sense that the occluding noisy patterns are not realistic. For that matter, and because there is currently no FER database that includes annotated partial occlusion ground truth, we also present on Figure \ref{qualitative} qualitative results on more realistic occlusions. Notice how the autoencoder network (learnt on CK+) assign high confidences (green) to non-occluded feature points, whereas examples that lie further from the captured manifold (e.g. because of lighting conditions, self-occlusion with a hand or with an accessory) are given lower values (red). The corresponding triangles are thus downweighted for FER and appear transparent on the last row. Also note that different facial regions can vote for different expressions, as shown on the second column (\textit{happy}+\textit{angry}/\textit{disgust}).

\subsection{AU detection}\label{AUsexp}

\subsubsection{Merging multiple datasets} In this section we present results for AU detection using LEP features. Table \ref{AUCCK} shows comparison of AUC for the prediction of AU activations on CK+ database obtained with LEPs trained on CK+, BU4D and SFEW databases, as well as models obtained \textit{via} the $M1$ and $M2$ strategies.

\begin{table*}[t]
\centering
\caption{AUC scores on CK+, BP4D and DISFA databases}
\label{AUCCK}
\scalebox{0.7}{
\begin{tabular}{ l | c | c | c | c | c || c | c | c | c | c || c | c | c | c | r }
&\multicolumn{5}{c}{CK+}&\multicolumn{5}{c}{BP4D}&\multicolumn{5}{c}{DISFA}\\
\hline
AU	& M1 & M2	& CK+	& BU4D &	SFEW	& M1 & M2	& CK+	& BU4D &	SFEW	& M1 & M2	& CK+	& BU4D &	SFEW\\
\hline
AU1	& 97.9 &	\textbf{98.4}	& 98.4	& 94.7	& 93.3& 59.6 & 62.7 &\textbf{63.6} & 60.9 &52.0 &66.1&68.4&\textbf{71.3}&57.6&66.6\\
AU2	& 98	& \textbf{98.2}	& 97.7	& 97.5	& 97.2 & 65.4 & 64.8& 62.3 & \textbf{66.0} &53.0&53.8&55.2&\textbf{67.3}&59.3&59.4\\
AU4	& 93.3	& \textbf{95.4}	& 94.8	& 83.1	& 85.6 &\textbf{68.7}&63.8&64.4&64.4&55.3&66.7&66.7&\textbf{67.3}&64.0&67.6\\
AU5	& 94	& \textbf{97.5}	& 95.5 &	93.2 &	95&-&-&-&-&-&84.2&85.6&73.3&\textbf{88.6}&73.7\\
AU6	& 95.4	& \textbf{95.7}	& 95.5 & 94.3 & 94.9&\textbf{83.1}&81.8&82.6&78.5&77.1&89.1&86.0&\textbf{89.2}&86.8&85.1\\
AU7	& 89.1 &	\textbf{90.2}	& 89.6	& 88.1 &	83&\textbf{76.8}&75.0&73.6&72.6&65.0&-&-&-&-&-\\
AU9	& 97.9	& \textbf{99.3}	& 98.7	& 98.5	& 94.8&-&-&-&-&-&\textbf{79.0}&77.0&75.4&74.0&53.4\\
AU10	& 83.7	& 85.6	& \textbf{86.5} &	78.4	& 81.7&83.7&\textbf{83.8}&83.3&81.0&78.6&-&-&-&-&-\\
AU12	& \textbf{97.6}	& 96	& 96.2	& 96	& 96.5&89.9&\textbf{90.0}&89.8&88.0&87.2&\textbf{95.5}&92.9&93.6&92.8&91.8\\
AU14	& -	& -	& -	& -	& -&65.2&66.4&63.7&\textbf{66.5}&64.9&-&-&-&-&-\\
AU15	& \textbf{91}	& 88.9	& 88.3	& 79	& 79.5&56.8&58.4&\textbf{58.5}&57.7&56.0&\textbf{69.5}&64.5&63.6&68.8&61.7\\
AU17	& 93.9	& \textbf{95.1}	& 93.4 &	81.5	& 86.4&55.8&65.7&\textbf{68.9}&65.1&60.6&\textbf{67.8}&61.2&53.5&59.1&58.8\\
AU20	& 91.9	& 93.8	& \textbf{94.5}	& 88.5	& 85.8&-&-&-&-&-&\textbf{65.0}&58.5&50.2&55.5&61.9\\
AU23	& -	& -	& -	& -	& -&50.1&57.2&\textbf{60.2}&57.5&54.2&-&-&-&-&-\\
AU24	& -	& -	& -	& -	& -&69.6&77.4&\textbf{78.2}&77.7&68.4&-&-&-&-&-\\
AU25	& 99	& \textbf{99.1}	& 98.8	& 87.1	& 97.4&-&-&-&-&-&94.8&95.0&94.0&\textbf{95.6}&80.0\\
AU26	& 75.7	& \textbf{81.2}	& 79.7	& 74.9	& 73.4&-&-&-&-&-&79.3&\textbf{81.4}&75.6&78.5&71.5\\
\hline
Avg	& 92.7	& \textbf{93.7}	& 93.4	& 88.2	& 88.9&68.8&70.6&\textbf{70.8}&69.6&64.3&\textbf{75.9}&74.4&72.9&73.4&69.3\\
\hline
\end{tabular}}
\end{table*}

For nearly every AU on CK+, the best AUC score is provided by the $M2$ strategy. However LEPs trained on CK+ only as well as the $M1$ strategy also provide good prediction results. LEPs trained solely on BU4D and SFEW seems a bit lackluster, but using the additional categorical expression data in addition to data from CK+ can be beneficial for prediction accuracy. Interestingly, on BP4D, LEPs trained on CK+ only seem to have a slight edge over the two LEPs models trained using all the available data. However, the $M2$ strategy and, to a lesser extent, $M1$ and training on BU4D only, provide close performances. Furthermore, on the DISFA dataset, the $M1$ and the $M2$ LEPs models provide the highest AUC. Overall, the $M2$ and $M1$ models seem to perform better, followed by the models trained on CK+. This proves that AU detection can benefit from additional training data labelled with categorical expressions. Finally, LEPs trained on SFEW did not perform very well, probably due to the fact that the database embraces too much variability for too few training data. Thus, the categorical expressions can not be captured adequately, as can be seen from the low accuracies showed in Section \ref{clean}.

\subsubsection{Comparison with state-of-the-art approaches} Table \ref{AUall} reports the overall best AUC obtained on the three datasets. It also draws a comparison between the AUC scores obtained using our method and results reported in recent publications involving similar protocols (same databases and sets of AUs, same intensity threshold for AU occurrence on DISFA). Our approach provides better results than SHTL \cite{ruiz2015shtl} on CK+, as well as accuracy similar to the multi-label CNN introduced in \cite{ghosh2015multi} on DISFA. Furthermore, it provides better performance than baselines LBP-TOP features used in \cite{zhang2014bp4d} on BP4D. This demonstrates that even though the AU detection pipeline is not meant to be optimal, LEPs learned on large amounts of categorical expression data yield high discriminative power for AU detection tasks. As such, an interesting direction would be to take into account the correlations between the tasks within a multi-output framework.

\begin{table}[t]
\centering
\caption{Comparison with other works}
\label{AUall}
\begin{tabular}{ l | c | r }
Database &	AUC(ours) & AUC(Other works)\\
\hline
CK+(14AU) & \textbf{93.7} & 91.7 (SHTL \cite{ruiz2015shtl})\\
\hline
BP4D(12AU) & \textbf{70.8} & 68.9 (LBP-TOP \cite{zhang2014bp4d})\\
\hline
DISFA(12AU) & \textbf{75.9} & 75.7 (Multi-label CNN \cite{ghosh2015multi})\\
\hline
\end{tabular}
\end{table}

\subsubsection{Relevance of AU confidence assessment} 
\begin{figure*}[htbf]
\centering
\includegraphics[width=\textwidth]{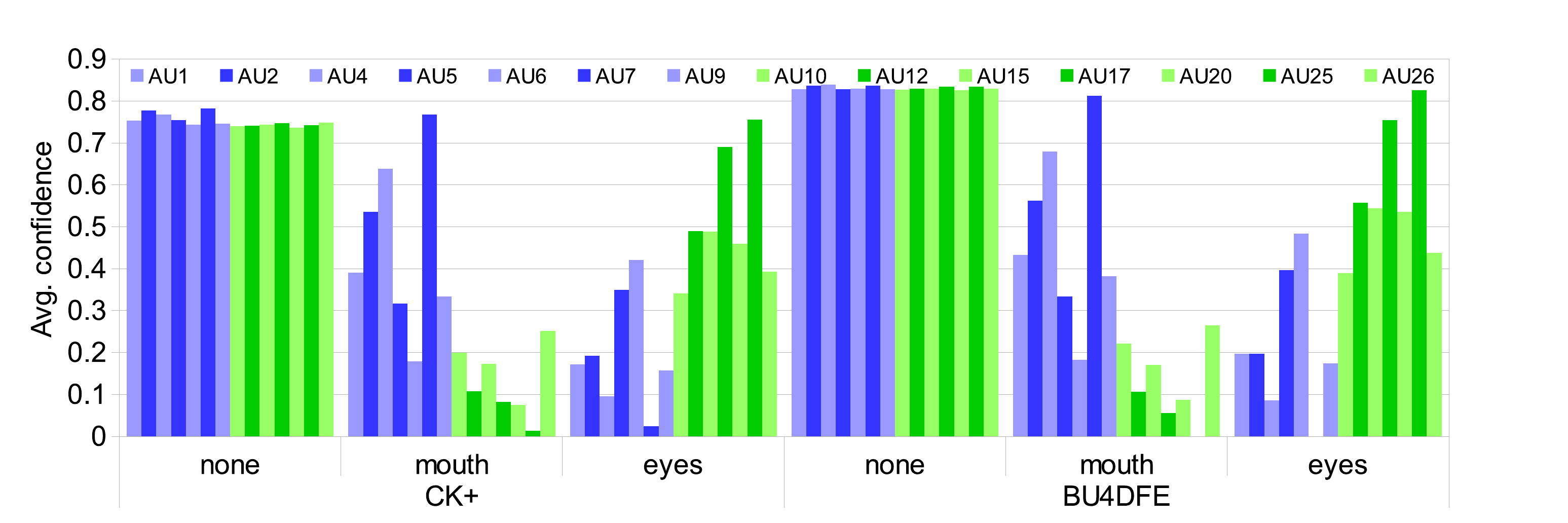}
\caption{AU confidence scores outputted on occluded CK+ and BU4D database}
\label{AUconfidencemes}
\end{figure*}

In order to assess the relevance of the AU-specific confidence measurement, we evaluated its average value on the occluded versions of the CK+ and BU4D databases generated in Section \ref{corr} for occlusion handling in categorical FER. From a general perspective, as can be seen on Figure \ref{AUconfidencemes}, low confidence measurements can be observed for AUs from the upper face region on the two scenarios involving eye occlusion. The same holds for AUs from the lower face region and the ``mouth occluded'' scenario, whereas the confidence scores are significantly higher in the non-occluded case. Interestingly, confidence scores for AU6 (cheek raiser) and, to a lesser extent, AU9 (nose wrinkle), are quite low even in the ``mouth occluded'' case. Indeed, as can be witnessed on Figure \ref{pautoau}, the confidence measurement for these AUs also use LEP features from the nose and mouth area.

\subsection{Computational load}\label{realtime}

The proposed framework for occlusion-robust FER (WLS-RF) and AU detection operates in real-time on video streams, even with large tree collections. Table \ref{evalTime} displays the elapsed time for each step of the evaluation pipeline. The test was performed on an Intel Core I7-4770 CPU on a single-thread C++/OpenCV implementation.

\begin{table}[t]
\centering
\caption{Measured evaluation time per processing step (in milliseconds)}
\label{evalTime}
\begin{tabular}{ l | r }
Processing step &	time (ms)\\
\hline
Feature point alignment & 10\\
\hline
Integral channels computation & 2\\
\hline
Confidence weights computation & 11\\
\hline
LEP computation (1000 trees)& 7\\
\hline
12 AU detection (50 trees) & 1\\
\hline
\textbf{Total} & \textbf{31}\\
\hline
\end{tabular}
\end{table}

It appears that the feature point alignment and confidence weight generation steps are the bottleneck of the system in term of computational load. However the computational load for the former can be reduced by the use of more efficient alignment algorithms such as the one in \cite{ren2014face}. As for the confidence weights, the computation time can be significantly reduced by a proper multithreading (e.g. computing the confidence for each feature point in parallel). As it is, the framework already runs at more than 30 fps even with large collections of trees. As for training, learning LEPs with 1000 trees on a big database (BU4D containing more than 8000 face images) took approximately three hours without parallelization. Training the hierarchical autoencoder network took half a day and learning the 12 AU detectors on DISFA database with 50 trees required one hour on the same I7-4770 CPU using a loose C++ implementation. Thus, our approach scales well both in terms of training and testing times, especially when compared to recent deep learning algorithms \cite{ghosh2015multi} for feature representation and learning.

\section{Conclusion and perspectives}\label{concl}

In this paper, we proposed a new high-level expression-driven LEP representation. LEPs are obtained from training Random Forests upon spatially defined local subspaces of the face. Extensive experiments on multiple datasets highlight the fact that the proposed representation improves the state-of-the-art for categorical FER and yields useful descriptive power for AU occurrence prediction. Furthermore, we introduced a hierarchical autoencoder network to model the manifold around specific facial feature points. We showed that the provided reconstruction error could effectively be used as a confidence measurement to weight the prediction outputted by the local trees. The proposed WLS-RF framework significantly adds robustness to partial face occlusions.

The ideas introduced in this work open a lot of interesting directions for future works on face analysis. First, note that the confidence weights are representative of the spatially defined local manifold of the training data. Thus, these confidence values can be used to determine which parts of the face are the most reliable in a general way (e.g. to address unpredicted illumination patterns or head pose variations), and are not limited to occlusion handling. Furthermore, we could inject confidence weights into the feature point alignment framework \cite{xiong2013supervised} to enhance the robustness of the feature point alignment w.r.t. occlusions. Compared to a discriminative approach using synthetic data \cite{ghiasi2014occlusion}, our manifold learning approach could in theory more efficiently deal with realistic occlusions. Moreover, the applications of LEPs for AU detection and intensity estimation are multiple. First, it would be interesting to learn LEPs using more expression data such as the datasets introduced in \cite{savran2008bosphorus,wallhoff2006database}, possibly with a more complex integration strategy. Also, it could be interesting to investigate the impact of using a more fine-grained facial mesh for FER and AU detection or intensity estimation using LEP representation, as it was done in \cite{jeni2015dense} for dense facial feature point alignment. Last but not least, the idea of learning Random Forests upon spatially defined local subspaces instead of random subspaces is not limited to face analysis, and could theoretically be applied in other fields such as gesture recognition.

\bibliographystyle{unsrt}
\bibliography{Biblio}   

\end{document}